%% file: main.tex
\documentclass[11pt]{article}

\usepackage[]{acl}

\usepackage{times}
\usepackage{latexsym}

\usepackage[T1]{fontenc}

\usepackage[utf8]{inputenc}

\usepackage{microtype}

\usepackage{inconsolata}

\usepackage{graphicx}
\usepackage{natbib}  

\usepackage{booktabs}
\usepackage{tabularx}
\usepackage{caption}
\usepackage{amsfonts}
\usepackage{amsmath}
\usepackage{color,xcolor} 
\usepackage{xspace}
\usepackage{adjustbox}
\usepackage{multirow}

\usepackage{amssymb}
\usepackage{algpseudocode}
\usepackage{xcolor,colortbl}
\usepackage{algorithm}
\usepackage{dcolumn}
\usepackage{longtable}
\usepackage{authblk}

\usepackage{graphicx}
\title{%
    \raisebox{-1em}{\includegraphics[height=3em]{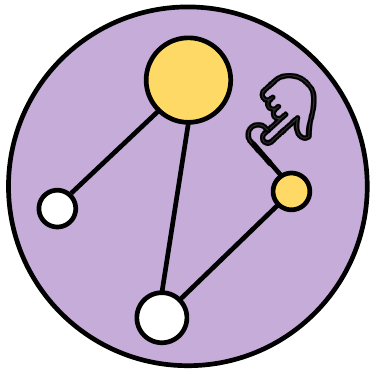}}%
    \hspace{0.3em}%
    XISM: an eXploratory and Interactive Graph Tool to Visualize and Evaluate Semantic Map Models%
}

\author{
  Zhu Liu\textsuperscript{*1},
  Zhen Hu\textsuperscript{*1},
  Lei Dai\textsuperscript{1},
  Yuxuan Liu\textsuperscript{1,2},
  Ying Liu\textsuperscript{1} \\
  \textsuperscript{1}School of Humanities, Tsinghua University, Beijing, China \\
  \textsuperscript{2}School of Environment and Society, Institute of Science Tokyo, Tokyo, Japan  \\
  \texttt{\{liuzhu22, huz25, yx-liu23\}@mails.tsinghua.edu.cn} \\
  \texttt{custdailei@126.com, yingliu@tsinghua.edu.cn}
}

\begin{document}
\maketitle
\renewcommand{\thefootnote}{\fnsymbol{footnote}}
\footnotetext[1]{Equal Contribution.}
\setcounter{footnote}{0} 
\renewcommand{\thefootnote}{\arabic{footnote}}
\begin{abstract}

Semantic map models visualize systematic relations among semantic functions through graph structures and are widely used in linguistic typology. However, existing construction methods either depend on labor-intensive expert reasoning or on fully automated systems lacking expert involvement, creating a tension between scalability and interpretability.
We introduce \textbf{XISM}, an interactive system that combines data-driven inference with expert knowledge. XISM generates candidate maps via a top-down procedure and allows users to iteratively refine edges in a visual interface, with real-time metric feedback. 
Evaluations of multiple cases and user surveys show that XISM improves linguistic decision transparency and controllability in semantic-map construction while maintaining scalability and computational efficiency.
The XISM system\footnote{\url{https://app.xism2025.xin/}}
, source code\footnote{\url{https://github.com/hank317/XISM}}
, and demonstration video\footnote{\url{https://youtu.be/m5laLhGn6Ys}} are publicly available.

\end{abstract}

\input{main/1_intro}

\input{main/2_related_work}

\input{main/4_system_design}

\input{main/5_user_case}

\input{main/6_evaluation_result}

\input{main/7_conclusion}


\bibliography{custom}

\appendix

\input{main/Appendix}

\label{sec:appendix}

\end{document}

%% file: main/1_intro.tex
\section{Introduction}



Semantic map models (SMMs) construct graphs that represent semantic or distributional associations among interrelated meanings or functions across cross-linguistic forms~\cite{haspelmath2003geometry,croft2003typology}. 
SMMs account for multifunctionality—where a single form encodes multiple functions—and support the identification of language-independent universal patterns, capturing a shared conceptual space across languages\footnote{In this paper, we use the terms ``meaning'', ``function'', and ``concept'' interchangeably.}. Through cross-linguistic comparisons, particularly focusing on function words that pose challenges for language learners, these models contribute to second language acquisition~\cite{linsemantics}, lexicography~\cite{RakhilinaRyzhovaBadryzlova}, and AI-driven educational applications~\cite{ye-etal-2025-position}.

The classical construction of SMMs adheres to the \textit{connectivity hypothesis}~\cite{haspelmath2003geometry}, which requires that all functions expressed by a given form must be connected within the graph. At the same time, the overall number of edges should be minimized in accordance with the \textit{economic principle}~\cite{georgakopoulos2018semantic}. To satisfy both constraints, linguists typically adopt an intuitive, bottom-up, and \textbf{knowledge-driven} procedure: they iteratively connect functions associated with each form and select an optimal configuration from multiple candidate graphs through linguistic reasoning.
However, this manual workflow is time-consuming and inefficient. Experts frequently need to revisit and revise earlier decisions to eliminate redundant edges, while the space of possible graph configurations grows combinatorially. These challenges severely restrict the scalability of SMM construction when working with larger datasets involving more forms, functions, and languages.

To overcome the limitations of manual construction, several algorithms have been developed to automate the building of SMMs. Moving beyond the traditional knowledge-driven paradigm, researchers have proposed top-down, \textbf{data-driven} methods grounded in graph theory~\cite{liu2025top, chenchen}. This approach begins by constructing a fully connected graph whose edge weights are determined by the cross-linguistic co-occurrence frequencies of functions. A spanning tree is then extracted under a reformulated ``maximum connectivity hypothesis'' together with metric-based constraints.
Unlike manual approaches, this method relies exclusively on multilingual data, requiring neither linguistic expertise nor iterative form-by-form construction. Consequently, it scales efficiently to large typological datasets. Nevertheless, its independence from linguistic input comes at the cost of interpretability and may overlook typologically meaningful structures.


To build a more user-friendly SMM, we adopt the view that SMMs should be \textbf{both knowledge-driven and data-driven}. After the automatic graph is constructed by the data-driven algorithm, experts can revise or refine edges based on typological and semantic knowledge. This \textit{human-in-the-loop} paradigm achieves a balance between computational efficiency and linguistic interpretability. 
To support such collaboration between humans and machines, an editable visualization interface equipped with evaluation mechanisms is also necessary, enabling researchers to adjust the structure interactively and assess the impact of modifications.

In this demonstration paper, we introduce \textbf{XISM}, an exploratory and interactive tool designed to \textbf{generate, visualize, edit, and evaluate} SMMs from user-uploaded datasets. The system begins by producing multiple candidate graphs derived from a fully connected network, guided by topological preferences, and presents them together with quantitative evaluation metrics. Users may highlight subgraphs associated with selected forms for detailed inspection and comparison. 
Building on our previous tree-based algorithm~\cite{liu2025top}, XISM further incorporates an efficient ``merge'' operation that adds necessary edges to ensure that all forms satisfy the connectivity hypothesis (i.e., achieving 100\% coverage).


After this initial construction, users may \textbf{edit} the graph to better align with linguistic knowledge and theoretical expectations. Evaluation metrics are updated in real time to support informed refinement, and the finalized graph can be exported as the optimal solution. 
We assess the effectiveness of XISM through offline experiments on typological datasets as well as online user studies. The results demonstrate that our human-in-the-loop framework is both efficient for model construction and effective for linguistically informed analysis.





Our contributions are threefold:

\begin{itemize}
    \item We introduce \textbf{XISM}, an interactive graph-based system that enables language typologists to construct, inspect, and refine SMMs within a human-in-the-loop framework.
    \item We extend our previous algorithm by incorporating an edge-merging mechanism and implementing a graph search strategy guided by typological evaluation metrics.
    \item We validate the system through offline experiments on typological datasets and online user studies, demonstrating that XISM is both efficient for model construction and effective for linguistically informed analysis.
\end{itemize}



%% file: main/2_related_work.tex
\section{Related Work}


\subsection{Semantic Map Models}

Semantic Map Models (SMMs) represent synchronic or diachronic relationships among meanings by constructing a graph based on cross-linguistic comparisons within a semantic domain. Compared forms include affixes~\cite{cysouw2011expression}, function words~\cite{zhang2017semantic}, content words~\cite{guo2012adjectives, cysouw2007building}, and constructions~\cite{malchukov2007ditransitive}. A single form may correspond to multiple meanings depending on context, exhibiting \emph{multifunctionality} for grammatical items~\cite{haspelmath2003geometry, zhang2015semantic} and \emph{colexification} for lexical items~\cite{franccois2008semantic}, with multifunctionality generally decreasing for larger linguistic units.

SMM construction follows two principles: the \textit{connectivity hypothesis}~\cite{haspelmath2003geometry}, requiring connected functions for each form, and the \textit{economic principle}~\cite{georgakopoulos2018semantic}, promoting sparsity. Two paradigms exist: \textbf{knowledge-driven}, relying on bottom-up, expert analysis~\cite{guo2010semantic,ma2015semantic_en}, and \textbf{data-driven}, which infers graphs from form-function data using algorithms~\cite{regier2013inferring, chenchen}. Our previous work~\cite{liu2025top} adopts a top-down, data-driven method that constructs a fully connected, co-occurrence-weighted graph and extracts a maximum-weight spanning tree, balancing connectivity and sparsity.


\subsection{Visualization Tools for SMMs}

Visualization is crucial in SMM research. Traditionally, semantic maps are manually drawn, with nodes as functions, edges as semantic associations, and connected areas as forms—consistent with a bottom-up construction paradigm. Early drafts are often visualized using general-purpose tools such as \textit{Gephi}, \textit{Cytoscape}, and \textit{Graphviz}~\cite{Bastian2009Gephi,Shannon2003Cytoscape,Gansner2000Graphviz}, but this typically involves separate construction and visualization stages.

Automatic, end-to-end visualization remains rare. CLICS~\cite{rzymski2020database} visualizes colexifications but does not follow classical SMM constraints, while dialect maps~\cite{montgomery2013geographic} capture geographic variation rather than abstract meaning relations. The closest comparable tool\footnote{\url{http://newlinguistics.cn/}} generates SMMs from user data but is static, non-interactive, and outdated. To our knowledge, \textbf{XISM} is the first interactive system supporting classical SMM construction with both automatic generation and expert refinement.

%% file: main/4_system_design.tex
\section{System Design}

\begin{figure}
    \centering
    \includegraphics[width=1.0\linewidth]{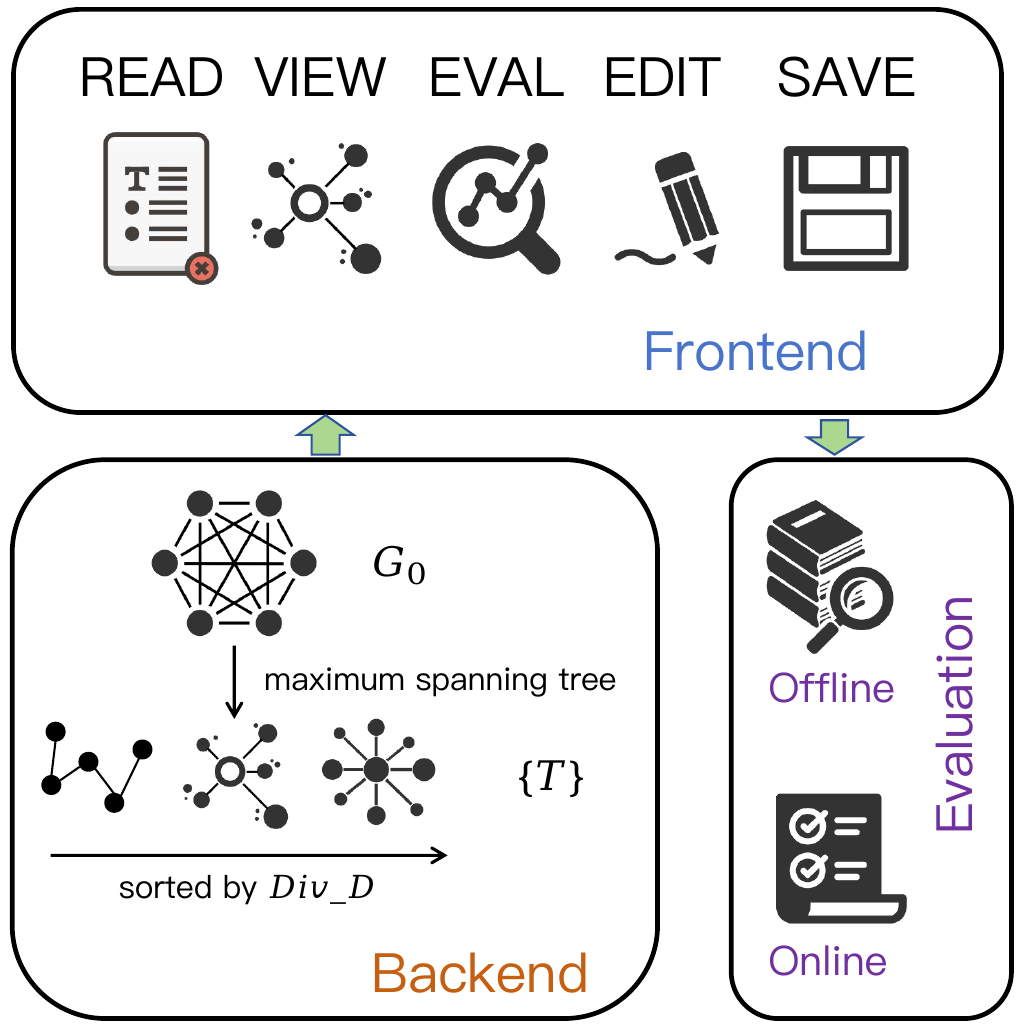}
    \caption{Overview of XISM}
    \label{fig:system}
\end{figure}

We design XISM as a web-based system with backend and frontend modules (Figure~\ref{fig:system}). The backend handles core algorithms including graph construction and dynamic evaluation, while the frontend manages data input/output, visualization of graph candidates and evaluations, form highlighting, and edge editing.

\subsection{Backend}

\subsubsection{Graph Construction} 
\label{sec:GC}


We build upon our previous algorithm~\cite{liu2025top} with key enhancements. Starting from a fully connected weighted graph $\mathcal{G}_0$, where edge weights denote function co-occurrence frequencies (i.e., the number of shared forms) and thus reflect semantic association strength, we extract a set of maximum spanning trees $\{\mathcal{T}\}$ as candidate networks. This ensures global connectivity, removes cycles to prevent uninterpretable semantic loops\footnote{A void graph lacks interpretability regarding the unidirectional trajectory of semantic evolution~\cite{haspelmath2003geometry}}, and maximizes the total edge weight. The resulting trees are ranked by degree standard deviation (Div\_D), with lower Div\_D empirically linked to better performance. 
%
%
\textcolor{black}{For efficiency, only the top $K$ candidates are retained, and $M$ of them are selected at regular intervals and passed to the frontend to promote diversity in the generated graphs.}


Since trees enforce acyclicity, some multifunctional forms may become disconnected. To address this, we iteratively merge remaning edges from $\mathcal{G}_0$ to reconnect them, selecting at each step one with the largest weight that maximally reduces the number of disconnected components. Details are in Appendix~\ref{app:merge}.

\begin{table}[h]
    \centering
    \small
    \begin{tabularx}{0.99\linewidth}{cXc} 
    \toprule
        Metric & Description & Trend \\
    \midrule
        Size & Total edge weight sum & $\leftrightarrow$ \\
        Cov & Proportion of covered instances & $\uparrow$ \\
        Avg\_D & Average node degree & $\leftrightarrow$ \\
        Div\_D & Standard deviation of node degrees & $\downarrow$ \\
    \midrule
    Acc & \textcolor{black}{Matched / total elements} in $\mathcal{G}$ & $\uparrow$ \\
        Prec & Matched edges / generated edges & $\uparrow$ \\
        Rec & Matched edges / annotated edges & $\uparrow$ \\
        F1 & Harmonic mean of Prec and Rec & $\uparrow$ \\
    \bottomrule
    \end{tabularx}
    \caption{Metrics for evaluating conceptual spaces. Arrows show preferred trend direction. Note that $\leftrightarrow$ indicates that a moderate value closer to human-annotated graph is preferred.}
    \label{tab:description}
\end{table}

\subsubsection{Graph Evaluation} 
Graphs are evaluated using metrics from Table~\ref{tab:description}, with unconnected forms reported for expert reference. Evaluations update dynamically upon frontend edits, enabling real-time assessment.

We revisit the correlation between Div\_D and accuracy within the candidate set $\{\mathcal{T}\}$, which is more meaningful given their high baseline performance. Additionally, we refine accuracy calculation by excluding diagonal elements (self-edges), which are always matched and do not affect evaluation.

\begin{figure*}[t]  
    \centering
    \includegraphics[width=\textwidth]{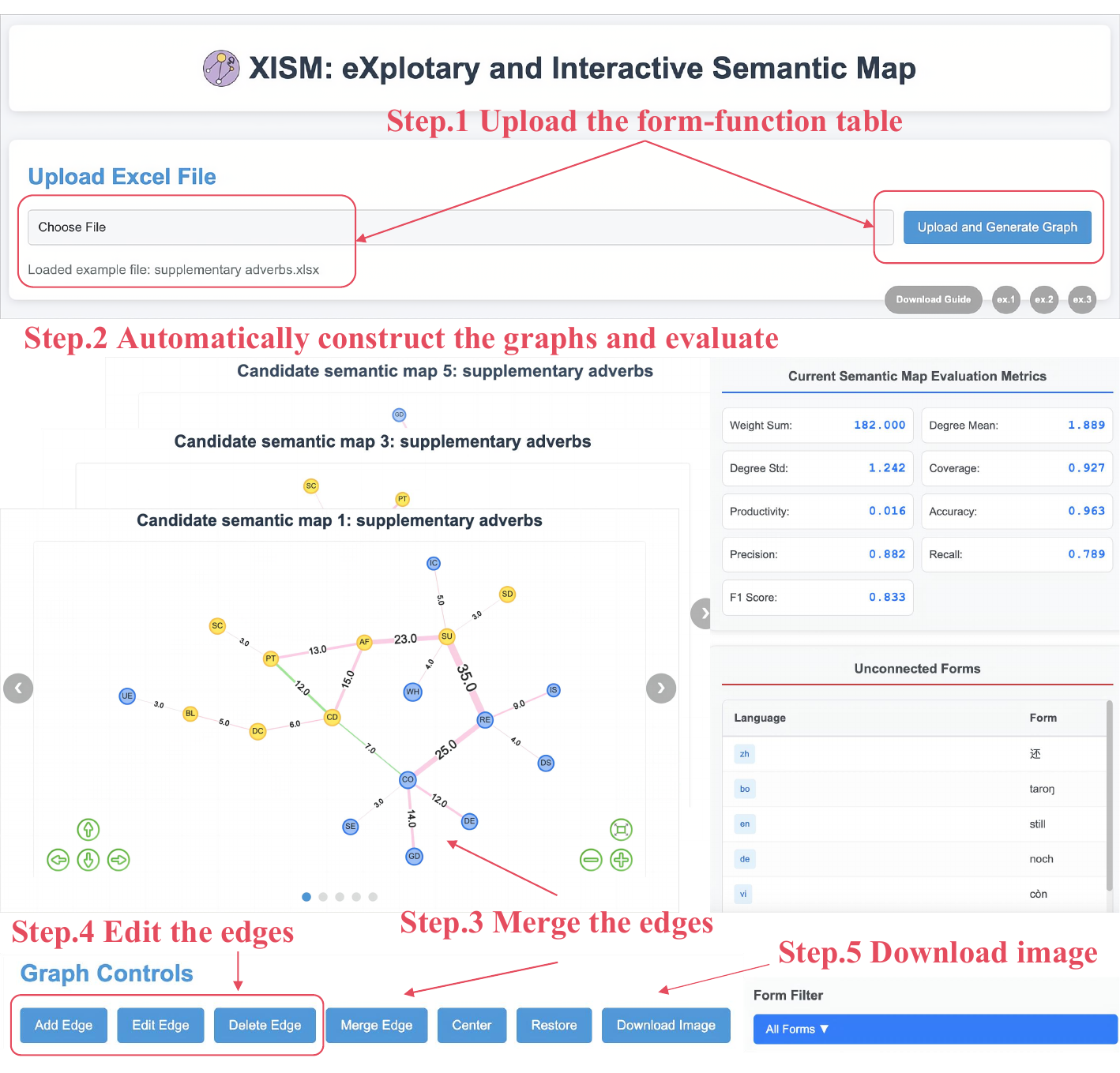}  
    \caption{XISM user interface and system walkthrough}
    \label{fig:system walkthrough}
\end{figure*}

\subsection{Frontend}
The user interface includes five modules: \textbf{READ}, \textbf{VIEW}, \textbf{EVAL}, \textbf{EDIT}, and \textbf{SAVE}. \textbf{READ} handles input of the binary form-function table $T$, which must have “languages” and “forms” as the first two columns; subsequent columns define user-specific functions. \textbf{VIEW} displays $M=5$ candidate graphs, with nodes as functions and edges weighted by association degree from $\mathcal{G}_0$—edge thickness reflects weight. Users can zoom, drag, center, and customize the visualization. A filter allows users to select a specific form and highlight its local graph, supporting the form-centered perspective commonly adopted in linguistic typology. The local subgraph for a specific form from a Filter can be highlighted, which is important for a form perspective as in linguistic typology. In \textbf{EDIT}, users can modify edge weights, add or delete edges, and merge edges to achieve 100\% coverage as described in Section~\ref{sec:GC}. Finally, \textbf{SAVE} allows exporting the final graph as a “.png” image.

\subsection{System Walkthrough}
Figure~\ref{fig:system walkthrough} outlines the user workflow:
\begin{enumerate}
    \item Upload the form-function table.
    \item Automatically construct and evaluate the graph.
    \item Merge edges to ensure full coverage.
    \item Edit edges based on linguistic knowledge.
    \item Save the finalized graph.
\end{enumerate}

%% file: main/5_user_case.tex
\section{System Evaluation Settings}
We evaluate XISM under both offline and online settings. Offline evaluation leverages established SMM cases from the literature, comparing our algorithm with the large language model (LLM), while online evaluation collects feedback from linguistic experts on various aspects.

\subsection{Offline Evaluation}


We collected 10 well-established SMM cases from the literature, each centered on a core semantic domain with its associated meanings or functions. The topics and scales of these cases vary substantially, providing a diverse set of SMMs for evaluation. Detailed descriptions and statistics of the cases are presented in Appendix~\ref{app:example}. 


We implement an LLM-based autonomous agent using the closed-source model Qwen3-Max\footnote{\url{https://qwen.ai/}} as a reference baseline. The agent is equipped with domain-specific knowledge of SMM construction through a carefully designed prompt\footnote{\url{https://juniperliuzhu.netlify.app/projects/smm_prompt_by_llm.html}}.Given a semantic domain, it first interprets the task instruction, then autonomously plans and executes the SMM generation process in multiple steps. The resulting graphs from this agent are compared against those generated by our method.


As for metrics, we measure the quality of generated SMMs both intrinsically and extrinsically~\footnote{Although most metrics follows our previous paper~\cite{liu2025top}, we change some ambiguous terms into a clearer one, and add more reasonable metrics when matching the reference, as specified in Appendix~\ref{app:metrics}.}. Intrinsic metrics include summed edge weights (Size), coverage (Cov) and degree statistics (mean Avg\_D and standard deviation Div\_D). Extrinsic evaluation compares $\mathcal{G}$ to expert-annotated gold standards via accuracy (Acc), precision (Prec), recall (Rec) and F1. Table~\ref{tab:description} summarizes these metrics; formulas are detailed in Appendix~\ref{app:metrics}.

\subsection{Online Evaluation}
We conducted an online survey to evaluate XISM’s utility, efficiency, usability, and overall impression, each rated on a 1–5 scale. The survey includes the three offline cases as examples but relies on user input for evaluation. Questionnaire details are in Appendix~\ref{app:surveyanswer}, which also contains five personal information questions. The target participants are SMM experts, though some users with limited experience were included to assess software usability.

%% file: main/6_evaluation_result.tex
\section{Evaluation Results}

We first analyze the efficiency of the XISM system by controlling the number of candidate trees, and then demonstrate its effectiveness through both offline and online evaluations.

\begin{figure}[h!]
    \centering
    \includegraphics[width=0.90\linewidth]{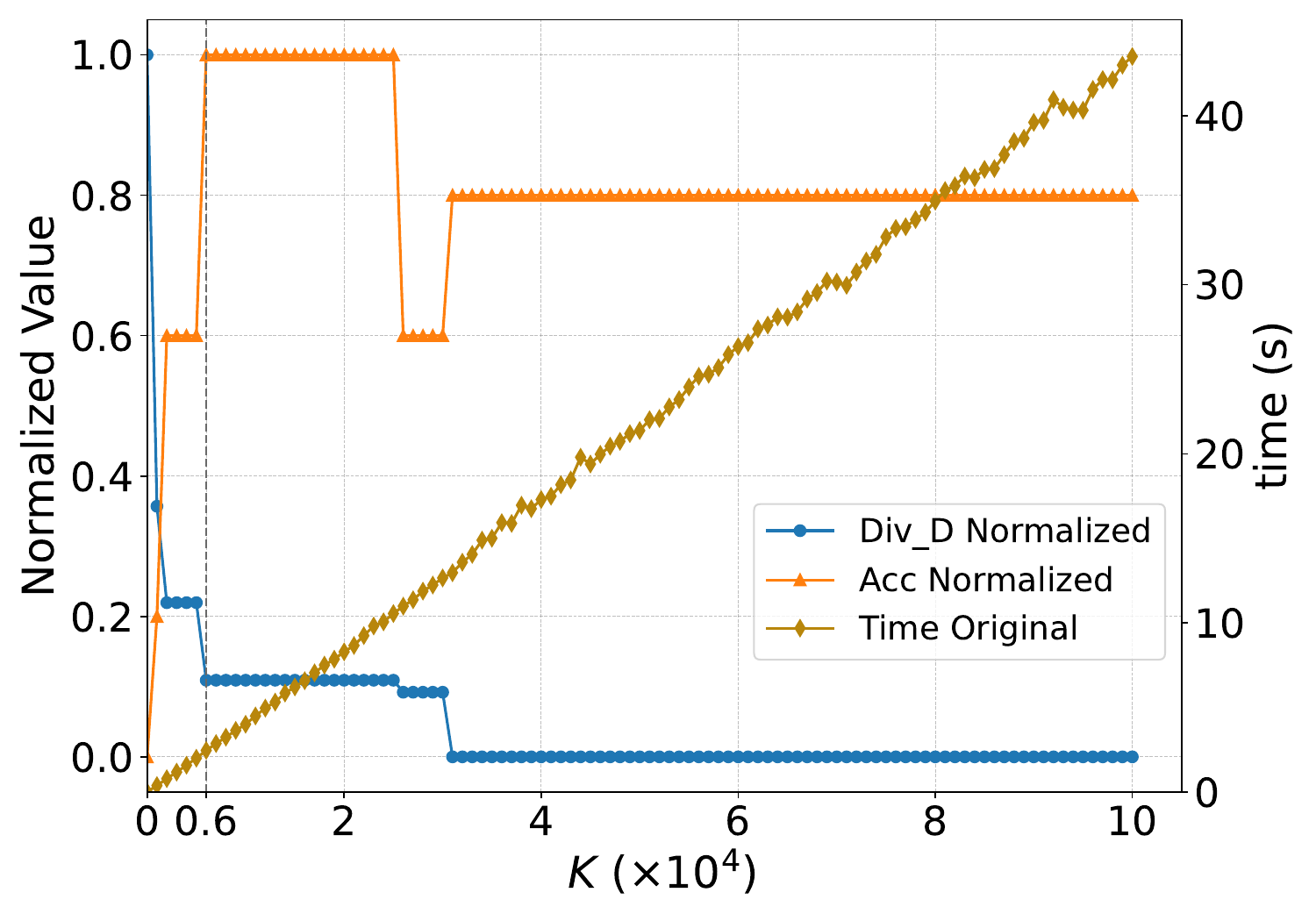}
    \caption{{Comparison of normalized Div\_D, Acc, and original computation time over different values of $K$.}}
    \label{fig:eff}
\end{figure}

\subsection{Efficiency}
The main computational tasks include two parts: generating the set of maximum spanning trees $\{\mathcal{T}\}$ and sorting these trees by their degree diversity (Div\_D) in ascending order. Following our previous work~\cite{liu2025top}, we employ Kruskal’s algorithm implemented in the Python package NetworkX\footnote{\url{https://networkx.org/}} to generate $\{\mathcal{T}\}$. This generation step is relatively fast, as shown in the results below. The computational bottleneck lies in sorting $\{\mathcal{T}\}$ by Div\_D, since the number of trees can reach hundreds of thousands as dataset size grows.

To improve efficiency, we approximate this sorting by selecting the top $K$ trees from $\{\mathcal{T}\}$ instead of sorting the full set. We experimentally determine the optimal $K$ on the EAT verbs case (the second case in Table~\ref{tab:exam_SUP}), which contains a huge space of candidate trees, as illustrated in Figure~\ref{fig:eff}. The figure shows that as $K$ increases, normalized Div\_D decreases, while both computation time and Acc increase; the Acc growth is notably nonlinear. The Pearson correlation between Acc and Div\_D is -0.515, indicating a moderate negative relationship.

Considering user experience, the largest tested $K$ results in over 10 seconds computation time, which is excessive for interactive use. We therefore select $K = 6,000$ as a compromise, where Acc saturates and the time cost remains reasonable. We then forward the 5 curated candidate graphs to the frontend for user interaction.

The lower part of Table~\ref{tab:other_metrics} shows the average time across ten cases to generate the final graph, comparing our two methods (Merged and Unmerged). Our approach achieves a substantial reduction in processing time—more than \textbf{100 times} faster than the LLM-generated baseline—demonstrating its high efficiency.

\begin{figure}[htbp]
    \centering
    \includegraphics[width=0.72\linewidth]{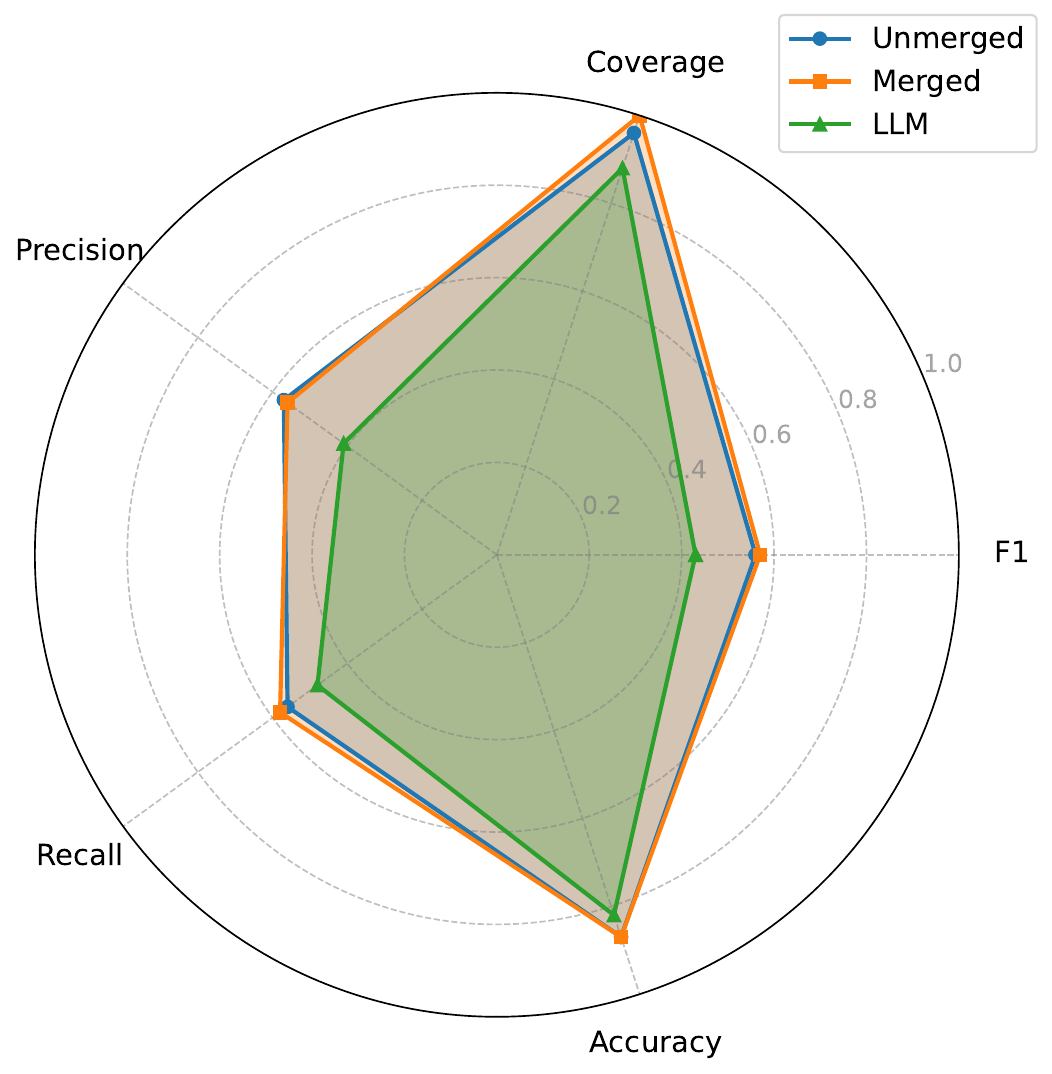}
    \caption{Extrinsic evaluation along with Coverage on our methods (Unmerged and Merged) and LLM. Higher values (or values closer to 1) indicate better performance.}
    \label{fig:radar}
\end{figure}

\subsection{Effectiveness}

\subsubsection{Offline Evaluation}



We compare our two algorithm versions of XISM, Unmerged and Merged, to LLM represented by Qwen. As shown in Figure~\ref{fig:radar}, XISM consistently outperforms the LLM-automated semantic map across all five evaluation dimensions, based on average~\footnote{Detailed evaluation results for each case are available on our website: \url{https://juniperliuzhu.netlify.app/projects/result_by_case}.} results over ten cases, demonstrating its ability to generate higher-quality semantic maps with improved structural fidelity and semantic completeness. Furthermore, the merged variant of XISM not only enhances connectivity compared to the unmerged version but also shows improvements in most other metrics.

\begin{table}[h!]
    \centering
    \small
    \begin{tabular}{c|ccc|c}
        \toprule
        Metrics & Unmerged & Merged & LLM & GT \\
        \midrule
        Size$\leftrightarrow$ & 186.50 & \textbf{200.20} & 237.30 & 204.30 \\
        \#Edges$\leftrightarrow$ & 15.60 & \textbf{16.70} & 23.10 & 16.82\\
        Div\_D$\downarrow$ & \textbf{1.40} & 1.48 & 2.00 & 1.472\\
        Avg\_D$\leftrightarrow$	& 1.83 & \textbf{1.94} &	2.58 & 1.97 \\ 
        \midrule
        Time(s)$\downarrow$ & \textbf{3.25} & 3.26 &	379.60  & - \\  
        \bottomrule
    \end{tabular}
    \caption{Other metrics and statistics for our methods (Unmerged and Merged), LLM and GT.}
    \label{tab:other_metrics}
\end{table}

We also compare additional statistics across our methods, the LLM baseline, and human-annotated ground-truth (GT), including number of edges (\#Edges), standard deviation of degree (\#Div\_D) and average degree (Avg\_D), and total time required to generate the final graph. As shown in Table~\ref{tab:other_metrics}, our methods yield results that more closely approximate the ground-truth (GT), while the LLM tends to generate more and denser edges, often exhibiting a star-like topology as reflected in its larger degree standard deviation.

\subsubsection{Online Evaluation}

\begin{figure}
    \centering
    \includegraphics[width=0.9\linewidth]{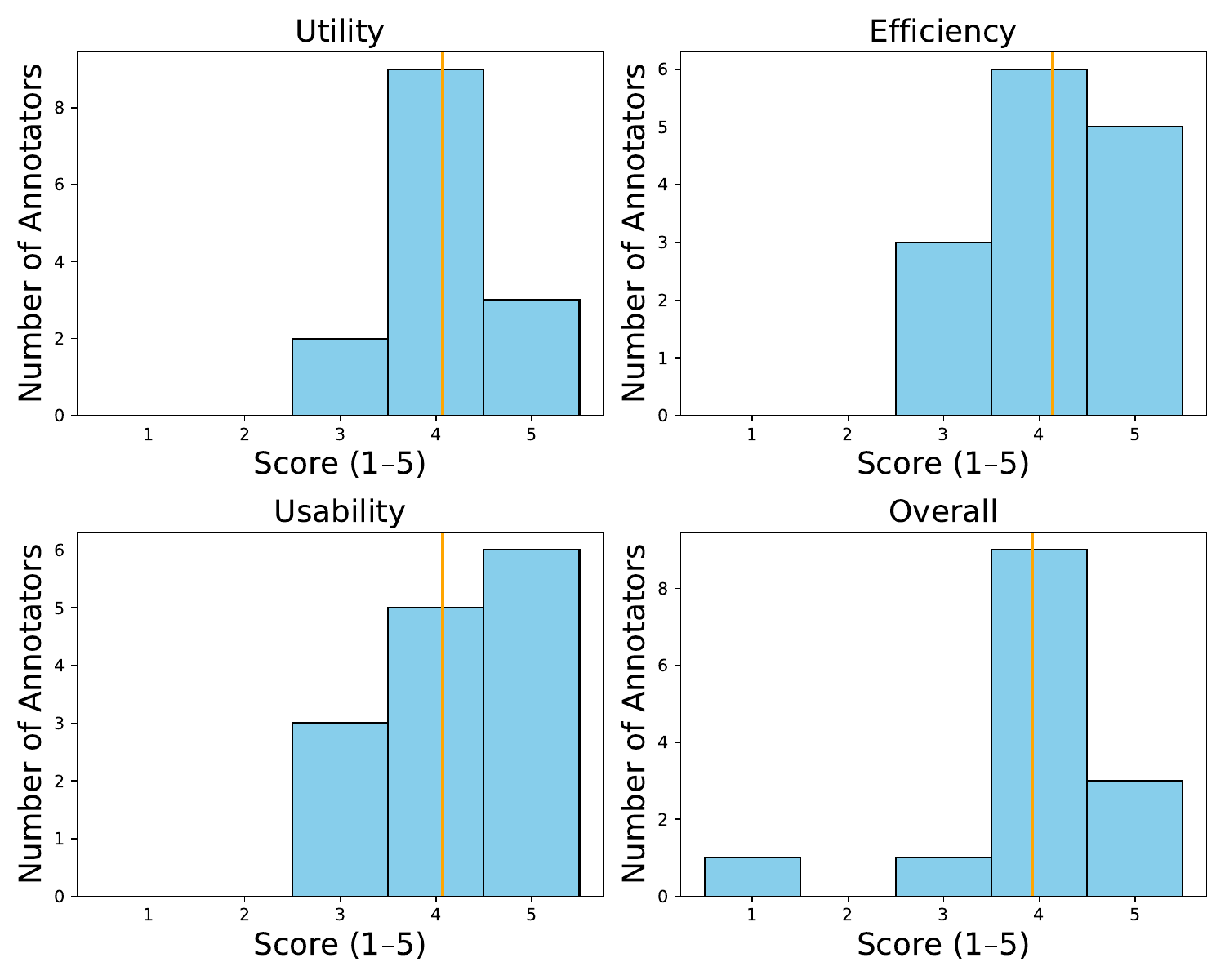}
    \caption{Survey summary from four dimensions. \textcolor{orange}{Orange} vertical line shows the average score.}
    \label{fig:survey}
\end{figure}

Figure~\ref{fig:survey} presents the score distributions for four dimensions: utility, efficiency, usability, and overall impression, corresponding to questions 6, 7, 8-9, and 14 in Appendix~\ref{app:surveyanswer}. Across all dimensions, XISM received favorable assessments, indicating its high effectiveness and efficiency from the users' perspectives.

%% file: main/7_conclusion.tex



\section{Conclusion}

In this demonstration paper, we present XISM, an exploratory and interactive graph tool designed for visualizing and evaluating semantic map models. Our system first generates candidate graphs from expert data and subsequently allows researchers to refine structures through edge editing with real-time metric feedback. This human-in-the-loop approach balances effectiveness and efficiency by integrating data-driven inference with knowledge-driven linguistic reasoning.

We anticipate that XISM will assist typologists and computational linguists in applying semantic map methodologies to larger datasets and more diverse semantic domains. Beyond that, the application will also contribute to second language or dialect acquisition by leveraging cross-linguistic conceptual comparison, within the broader context of AI for education. 




\section*{Ethics Statement and Limitations}
We ensure the confidentiality of personal information collected during XISM’s evaluation to protect user privacy. Furthermore, no user-uploaded data from the frontend is accessible to us unless they share with us. We only use generative AI to help us polish the manuscript.


We acknowledge several limitations of XISM. First, certain components—such as the sorting of maximum spanning trees—remain computationally inefficient when handling much larger-scale data. To address this, we plan to explore approximation algorithms to improve scalability and performance.

Second, although LLM serves as a baseline in our evaluation, the current approach relies solely on graph-theoretic methods without incorporating language models or agentic mechanisms. Future work will investigate integrating LLM-based semantic inference and autonomous agents to enhance the intelligence and automation of SMM construction.

Third, the current evaluation is limited to linguistic typology cases. We aim to extend XISM to broader application scenarios, particularly in AI for education, where semantic maps could support curriculum learning and conceptual modeling. Identifying and adapting to such domain-specific requirements will be a key direction for future development.

%% file: main/Appendix.tex
\clearpage
\section{Appendix}
\algnewcommand{\Sort}[1]{\State \textbf{sort} #1}

\subsection{Merging Algorithm}
\label{app:merge}

We present our ``merge'' pseudo-algorithm in Algorithm~\ref{alg:merge}. This function adds the necessary edges to guarantee 100\% coverage. \textcolor{black}{Starting from the initial graph with weighted edges, we first identify all disconnected components by traversing every form. For each disconnected form, we add iteratively edges in descending order—first by the number of disconnected components that contain edge $e$, then by weight.} After each addition, we update the set of disconnected components and the candidate edge list. This process repeats until the graph becomes fully connected for each form (100\% coverage).

\begin{algorithm}[H]
    \caption{Merge algorithm for full coverage}
    \label{alg:merge}
    \begin{algorithmic}[1]
    \Require A weighted graph $\mathcal{G}_0 = (V, E, w)$, a graph produced by unmerged XISM $\mathcal{T}$
    \Ensure A graph $\mathcal{T}'$ with 100\% coverage
    
    \State Identify all disconnected components $\mathcal{C}$ by traversing every form
    \State Identify the candidate edges $E_{\text{cand}}$ which belong to $\mathcal{C}$ but not to $\mathcal{T}$
    
    \State $\mathcal{T}' \gets \mathcal{T}$

    \While{$|\mathcal{C}| > 1$}
        \State Sort $e \in E_{\text{cand}}$ in descending order:
        \State \quad Primary key: number of unconnected components containing edge $e$
        \State \quad Secondary key: weight $w(e)$
        \State Select the top edge $e$ from sorted $E_{\text{cand}}$
        \State Add $e$ to $\mathcal{T}'$
        \State Update $\mathcal{C}$
        \State Update $E_{\text{cand}}$
    \EndWhile
    
    \State \Return $\mathcal{T}'$
    \end{algorithmic}
    \end{algorithm}

\begin{table*}[h!]
    \centering
    \small
    
    \begin{tabular}{c|c|ccc|c|ccc}
        \toprule
        Name & Source & \#L & \#F & \#Fc & Sparsity & Avg\_D & Div\_D & \#Edges \\
        \midrule
        Supplementary Adv. & \cite{guo2010semantic} & 36 & 82 & 18 & 0.816 & 2.111 & 1.560 & 19\\
        EAT verbs & \cite{deng2020semantic} & 42 & 42 & 17 & 0.648 & 1.882 & 1.711 & 16 \\
        Ditransitive & \cite{guo2013ditransitive} & 6 & 29 & 34 & 0.737 & 1.941 & 1.454 & 33 \\
        And \textit{He} & \cite{ma2015semantic_zhai} & 55 & 55 & 29 & 0.801 & 2.000 & 1.509 & 29 \\
        What \textit{Shenme} & \cite{noda2015semantic_shenme} & 25 & 44 & 20 & 0.556 & 2.900 & 2.142 & 29\\
        Quantifier & \cite{li2011cross} & 90 & 90 & 14 & 0.590 & 1.857 & 1.059 & 13 \\
        Make \textit{Zuo} & \cite{liu2024semantic} & 17 & 17 & 8 & 0.375 & 1.750 & 1.090 & 7\\
        Hit \textit{Da} & \cite{liu2024semantic} & 14 & 14 & 19 & 0.417 & 1.895 & 2.292 & 18 \\
        Come \textit{Lai} & \cite{liu2024semantic} & 14 & 14 & 8 & 0.420 & 1.750 & 1.392 & 7\\
        Tree/Wood/Forest & \cite{georgakopoulos2018semantic} & 4 & 13 & 5 & 0.692 & 1.600 & 0.800 & 4 \\
    \bottomrule
    \end{tabular}
    \caption{Well-established cases from typological literature}
    \label{tab:cases_info}
\end{table*}

\subsection{Data Examples}
\label{app:example}
We collected 10 well-established cases from the linguistic typological literature. 
Table~\ref{tab:cases_info} presents for each case: its name, source, scale (including number of languages \#L, forms \#F, and functions \#Fc), and graph statistics of the SMM annotation, including sparsity (proportion of zeros in the form-function table), average degree (Avg\_D), standard deviation of degree (Std\_D), number of edges (\#Edges), and size (summed weight).

\begin{table}[]
    \centering
    \begin{tabular}{cc|ccccc}
        \toprule
        language & form & AF & SU & RE & CO & GD \\
        \midrule
        Chinese & hai & 0 & 1 & 1 & 1 & 1 \\
        Chinese & you & 0 & 1 & 1 & 0 & 0 \\
        Chinese & ye & 1 & 1 & 0 & 0 & 0 \\
        Chinese & zai & 0 & 1 & 1 & 1 & 1 \\
        English & also & 1 & 1 & 0 & 0 & 0 \\
        English & too & 1 & 1 & 0 & 0 & 0 \\
        Engish & again & 0 & 1 & 1 &0 & 0 \\
        German & auch & 1 & 1 & 0 & 0 & 0 \\
        French & aussi & 1 & 1 & 0 & 0 & 0 \\
        Japanese & mo & 1 & 1 & 0 & 0 & 0 \\
        Italian & anche & 1 & 1 & 0 & 0 & 0 \\ 
        \bottomrule
    \end{tabular}
    \caption{Partial data from SUP. ``AF'', ``SU'', ``RE'', ``CO'' and ``GD'' are short of these functions: additive focus, supplement, repetition, continuation, greater degree.}
    \label{tab:exam_SUP}
\end{table}

The name reflects the core semantic domain under investigation. 
For example, \textit{Supplementary adverbs} indicates that all forms are adverbs and relate to the meaning of supplementation, such as \textit{again} and \textit{still} in English, and \textit{hái} and \textit{yòu} in Chinese. We extract partial forms and functions for this case in Table~\ref{tab:exam_SUP} as an illustration, and refer readers to the original studies for more details. 

Some cases are large-scale from a linguistic perspective—for instance, \textit{Supplementary adverbs} and Quantifier each involve nearly one hundred forms or languages. Our automated tool XISM can efficiently generate SMMs for such large-scale cases.

\subsection{Evaluation Metrics}
\label{app:metrics}

We define metrics to evaluate the conceptual space $G$ both intrinsically and extrinsically.

\subsubsection{Intrinsic Metrics}

\begin{itemize}
    \item \textbf{Size}: total edge weight: 
    \[
    \text{Size} = \sum_{(i, j) \in E} w_{ij}
    \]
    \item \textbf{Coverage}: proportion of connected forms $F'$ among observed forms $F$: 
    \[
    \text{Coverage} = \frac{|F'|}{|F|}
    \]
    \item \textbf{Avg\_D}: average node degree: 
    \[
    \text{Avg\_D} = \frac{1}{|V|} \sum_{v \in V} \deg(v)
    \]
    \item \textbf{Div\_D}: standard deviation of node degrees: 
    \[
    \text{Div\_D} = \sqrt{\frac{1}{|V|} \sum_{v \in V} (\deg(v) - \text{Avg\_D})^2}
    \]
\end{itemize}

\subsubsection{Extrinsic Metrics}

Let $G(w>0)$ denote edges with positive weight.  

\begin{itemize}
    \item \textbf{Accuracy (Acc)}: fraction of correctly predicted edges: 
    \[
    \text{Acc} = \frac{|\mathcal{G} \cap \mathcal{G}^\ast|}{|\mathcal{G}^\ast|}
    \]
    \item \textbf{Precision (Prec)}: correctly predicted edges over predicted edges: 
    \[
    \text{Prec} = \frac{|\mathcal{G}(w>0) \cap \mathcal{G}^\ast(w>0)|}{|\mathcal{G}(w>0)|}
    \]
    \item \textbf{Recall (Rec)}: correctly retrieved edges over gold edges: 
    \[
    \text{Rec} = \frac{|\mathcal{G}(w>0) \cap \mathcal{G}^\ast(w>0)|}{|\mathcal{G}^\ast(w>0)|}
    \]
    \item \textbf{F1-score (F1)}: harmonic mean of Precision and Recall: 
    \[
    \text{F1} = \frac{2 \cdot \text{Prec} \cdot \text{Rec}}{\text{Prec} + \text{Rec}}
    \]
\end{itemize}

\subsection{Survey Content for XISM}
\label{app:surveyanswer}

This questionnaire evaluates the interactive semantic map software \textbf{XISM}. Participants may use either their own data or the provided sample data. Responses are confidential and used for research purposes only.

Unless otherwise specified, Likert-scale questions use a 1--5 scale (1 = lowest evaluation, 5 = highest evaluation).

The evaluation focuses on four dimensions: \textit{utility}, \textit{efficiency}, \textit{usability}, and \textit{overall impression}, corresponding to Q6, Q7, Q8-9, and Q14, respectively.

\begin{enumerate}
    \item[Q1.] Your name.
    \item[Q2.] Your current position.
    \item[Q3.] Your email address.
    \item[Q4.] Your academic background / research area.
    \item[Q5.] Familiarity with semantic map models (1--5).
    \item[Q6.] Usefulness of the semantic maps generated by XISM for research (1--5).
    \item[Q7.] Generation efficiency of XISM (1--5).
    \item[Q8.] Overall usability of XISM (feature completeness and workflow smoothness) (1--5).
    \item[Q9.] Ease of use of XISM (interface intuitiveness and beginner-friendliness) (1--5).
    \item[Q10.] Whether you used your own data.
    \item[Q11.] Willingness to use XISM in future research.
    \item[Q12.] Willingness to recommend XISM to others.
    \item[Q13.] Interest in receiving future updates about XISM.
    \item[Q14.] Overall rating of the software (1--5).
    \item[Q15.] Problems encountered or suggestions for improvement.
\end{enumerate}